# Utilizing AI Language Models to Identify Prognostic Factors for Coronary Artery Disease: A Study in Mashhad Residents

**Bami Zahra[1], Behnampour Nasser[2]\*, Doosti Hassan[3], Ghayour Mobarhan Majid[4]**

1. Research Fellow, Center for Biostatistics, Epidemiology and Public Health (C-BEPH) Dept. of Clinical and Biological Sciences - University of Turin.
2. Assistant Professor of Biostatistics, Department of Biostatistics and Epidemiology, Health Management and Social Development Research Center, Golestan University of Medical Sciences, Gorgan, Iran
3. Assistant Professor of Biostatistics, School of Mathematical and Physical Sciences, Macquarie University, Sydney, NSW 2109, Australia
4. Professor of Nutritional Sciences, Department of Nutrition, School of Medicine, Metabolic Syndrome Research Center, Mashhad University of Medical Sciences, Mashhad, Iran.

**Abstract:**

**Background:**

Understanding cardiovascular artery disease risk factors, the leading global cause of mortality, is crucial for influencing its etiology, prevalence, and treatment. This study aims to evaluate prognostic markers for coronary artery disease in Mashhad using Naive Bayes, REP Tree, J48, CART, and CHAID algorithms.

**Methods:**

Using data from the 2009 MASHAD STUDY, prognostic factors for coronary artery disease were determined with Naive Bayes, REP Tree, J48, CART, CHAID, and Random Forest algorithms using R 3.5.3 and WEKA 3.9.4. Model efficiency was compared by sensitivity, specificity, and accuracy. Cases were patients with coronary artery disease; each had three controls (totally 940).

**Results:**

Prognostic factors for coronary artery disease in Mashhad residents varied by algorithm. CHAID identified age, myocardial infarction history, and hypertension. CART included depression score and physical activity. REP added education level and anxiety score. NB included diabetes and family history. J48 highlighted father's heart disease and weight loss. CHAID had the highest accuracy (0.80).

**Conclusion:**

Key prognostic factors for coronary artery disease in CART and CHAID models include age, myocardial infarction history, hypertension, depression score, physical activity, and BMI. NB, REP Tree, and J48 identified numerous factors. CHAID had the highest accuracy, sensitivity, and specificity. CART offers simpler interpretation, aiding physician and paramedic model selection based on specific.

**Introduction:**

Coronary artery disease [CAD] is caused by the development of atherosclerotic plaques on the epicardial coronary artery wall, which leads to the lumen of the artery's narrowing and obstruction of blood flow. With dietary modifications and medical treatment, CAD can be postponed or even averted (1).

Coronary artery disease organizes the largest percentage of cardiovascular diseases (2). One of the biggest challenges in coronary intervention has been managing CAD (3). The mortality rate of cardiovascular disease has fallen from 10 percent to 30 percent in the 20th century. It is predicted that more than 23.6 million deaths will occur in different communities in 2030 related to cardiovascular disease (4-7). The most common cause of death in the US is coronary artery disease [CAD], which is estimated to cause 610,000 deaths annually [1 in 4 deaths]. It is responsible for

17.8 million deaths per year and is the third biggest cause of mortality in the world. In the US, the yearly cost of healthcare services for CAD is expected to be more than $200 billion. CAD can be avoided, despite being a major cause of mortality and disability (8).

In Iran, coronary artery disease has been reported as the first and most common cause of death in all ages. As many as 317 deaths out of a total of 700 to 800 daily deaths are attributed to cardiovascular disease (9). The progress in medical science over the past few decades has made it possible to identify risk factors that may accelerate coronary artery disease, but this knowledge has not led to a significant reduction in the incidence of coronary heart disease, yet. Factors that accelerate coronary artery disease can be divided into two categories: modify and unmodified (10-12). Concerning the past research, unmodified risk factors, which cannot be altered through intervention as follows: age, gender, family history, and genetic traits, and modifiable risk factors that can change with lifestyle changes as follows: smoking status, hypertension, Diabetic, hyperlipidemia, high-density lipoprotein, low-density lipoprotein, and triglycerides (13-16). The purpose of the present study was done by identifying the risk factors of coronary artery disease and their affection in Mashhad to deal with the risk factors of this disease.

One of the most powerful and well-liked techniques in knowledge data mining and discovery is decision trees (17). A decision tree can be provided for different procedures in medicine to guide doctors towards the most reliable and predictable treatment approach based on structures. A decision tree for Alveolar ridge preservation procedures was offered by Steigmann et al. to help doctors choose the most conservative and predictable course of action based on the remaining socket anatomical components following extraction (18).

The decision tree is one of the nonparametric methods of classification that is divided into two categories. We use classification trees for classified dependent variables and regression trees for the continuous dependent variables. The classification tree is similar to discriminant function Analysis and logistic regression. In this method a set of logical conditions is used as an algorithm with a tree structure, to classify or predict an outcome. Nonlinearity and ease of interpretation of results are the two main advantages of using this method over many prediction models. Therefore, this model does not require the default linear relationship between the predictor variables and the outcome variable (19) . Various algorithms are presented with concerning factors and different methods for determining the decision tree. The most important and known algorithms are as follows: CHAID CART, ID3, C4,5, J48, NB, REP Tree, Random Forest [RF], and CRUISE QUEST.

**Materials and Methods:**

The data of this study are the result of a cohort study named MASHHAD-STUDY [Mashhad stroke and heart atherosclerotic disorder]. It started in 2009 in Mashhad. In this cohort study, 11,000 people were enrolled. 9761 individuals were diagnosed as healthy and remained in the study, after examining the conditions of the individuals according to the inclusion criteria. Then, demographic data, anthropometric measurements, lifestyle information including smoking habits, education level, Beck Anxiety Inventory, Beck Depression Inventory, and a checklist of factors affecting coronary artery disease were collected. All remaining individuals in the study were monitored for cardiovascular status, once during the years 2011–2014, and again, during the period 2015–2016 by telephone. After the obligatory examinations, only 235 of 768 people who considered themselves to be coronary artery disease by self-declaration, were confirmed and remained in the study (20). Therefore, in this study, the case group was a total of 235 individuals who were confirmed as coronary artery disease patients. Randomly, for each case, three controls were selected from 9,482 healthy non-cardiovascular patients. The first ones with incomplete information were excluded to select the control group, then, a random sample of 705 individuals was selected from the rest of the healthy subjects by using a select option in SPSS 25 software. After merging the case and controls, the file was transferred to R3.5.3 software and the appropriate variables were selected the result was reported by using CART and CHAID algorithms in the RATTLE package (21). Then RF, NB, REP Tree, and J48 algorithms were implemented and the results were reported by using WEKA 3.9.4 software. All the descriptive statistics were used by R 4.3.3. The variables used in this study included 29 independent variables and one dependent variable as follows: gender, age (aghe_1), degree, job status, marital status, physical activity level (PAlfinal), waist circumference(WHR_10), anxiety score (Ans), depression score (DepS), cholesterol levels, HDL levels, LDL levels, triglyceride levels, Diabetic, history of coronary artery disease of father, mother, brother and sister, history of hypertension, history of diabetes hypertension, increased blood lipids, osteoporosis, diet change, myocardial infarction, cigarette smoking, weight lost in recent trimester, body mass index (BMI), bone fractures and Coronary artery disease (CHD) as the independent variable. A decision tree is a method that uses various algorithms to display data in tree form and classify them into separate classes. In the early 1980s, Brayman coined the term decision tree (22). The main purpose of the decision tree is to detect the structural information that exists in the data. The tree is broken down into sub-branches according to a branch criterion, and this process is repeated periodically until the data for each group should be in the same category. To express the extracted rules, the path of the root to the leaf of the tree is scrolled and the rules are expressed in conditional form. The most common algorithms are as follows: ID3, CART, CHAID, C4.5 [J48], and RF.

**Results:**

Of 940 individuals studied, 393 [41.8%] were male and 547 [58.2%] were female. The age range was 33-71 years. 882 [93.8%] of them were married, 2 [0.2%] were single and 56 [6.0%] were divorced or widowed (as shown in Table 1).

| Variable | Definition | Count | Percentage |
|---|---|---|---|
| Gender | Male | 393 | 41.8 |
|  | Female | 547 | 58.2 |
| Marital status | Married | 882 | 93.8 |
|  | Divorced or widow | 56 | 6.0 |
|  | Single | 2 | 0.2 |
| Level of Education | Illiterate | 139 | 14.8 |
|  | Elementary | 384 | 40.9 |
|  | Associate | 308 | 32.8 |
|  | Bachelor's | 100 | 10.7 |
|  | Master's and doctoral degrees | 9 | 0.7 |

Table. 1: Frequency distribution of the subjects studied by gender, marital status, and education level.

The descriptive statistics, such as Minimum, Maximum, Mean, Q1, Q3, and standard deviation of 10 numeric variables are shown in Table 2.

| Variable | Mean | StDev | Minimum | Q1 | Q3 | Maximum |
|---|---|---|---|---|---|---|
| LdL | 117.57 | 34.28 | 42.22 | 94.90 | 137.61 | 280.80 |
| BMI | 28.108 | 4.890 | 16.222 | 24.806 | 31.168 | 46.557 |
| AnS | 11.033 | 10.249 | 0.000 | 3.000 | 16.000 | 62.000 |
| DepS | 12.556 | 9.888 | 0.000 | 5.000 | 18.000 | 63.000 |
| Cholesterol | 191.43 | 40.05 | 92.00 | 164.00 | 215.75 | 413.00 |
| HDL | 42.103 | 9.510 | 14.000 | 35.625 | 47.800 | 86.500 |
| TRiGli | 144.98 | 82.98 | 22.00 | 87.00 | 178.50 | 520.00 |
| PALfinal | 1.5770 | 0.2766 | 0.8411 | 1.3990 | 1.7435 | 2.6921 |
| age_1 | 49.560 | 8.469 | 33.000 | 43.000 | 56.000 | 71.000 |
| WHR_10 | 9.2707 | 0.8029 | 6.4957 | 8.7736 | 9.7811 | 16.5455 |

Table 2: The descriptive statistics of the numeric variable(s)

Fig 1 demonstrates the distribution of those variables with a normal curve that has been fitted onto the plots.

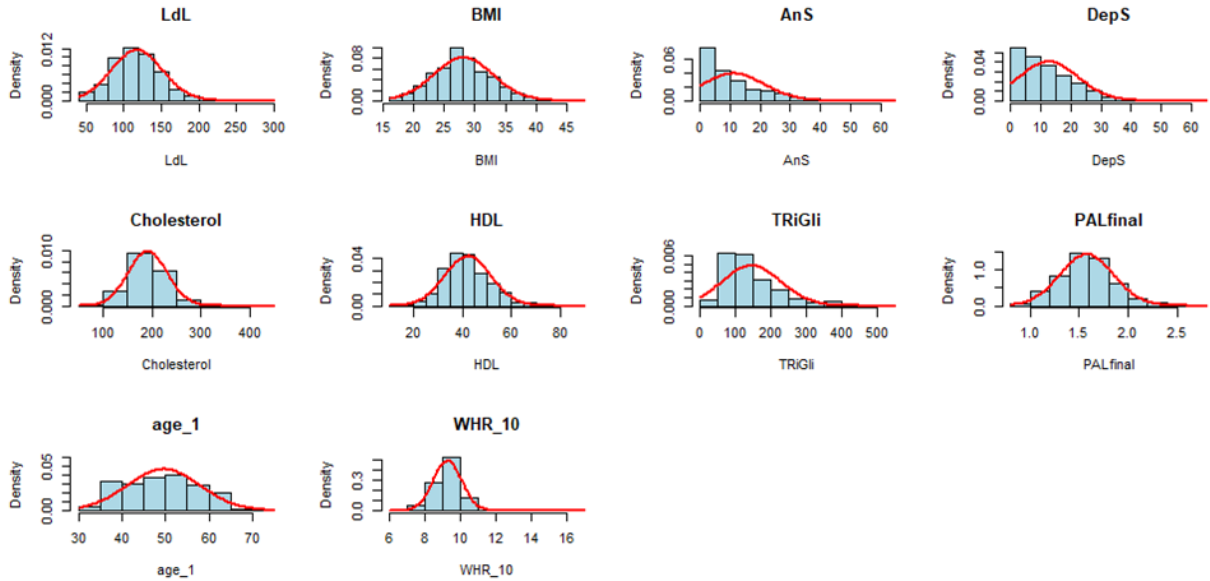

Fig 1: The distribution of numeric variables

To enhance data comprehension, we utilize a heatmap for both cases and controls, as depicted in Fig 2.

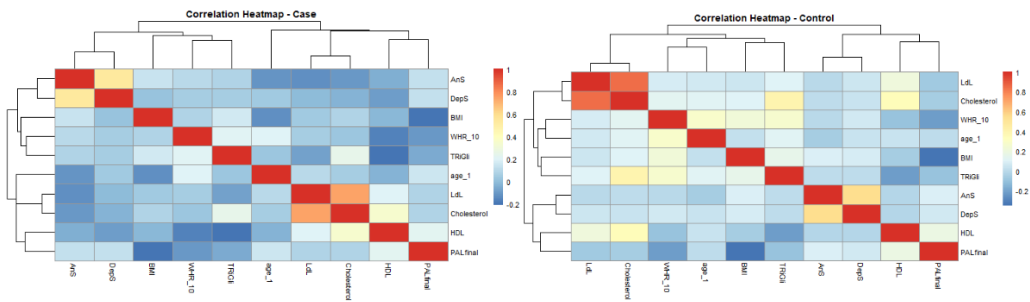

Fig 2: The heatmap illustrates the data for the case and control groups

In our previous study (23) , we identified several variables that influenced CVD through logistic regression, and in this study, we identified additional variables using various ML algorithms. Here, we present conditional box plots for these important variables in case and control groups. The results are depicted in Fig 3.

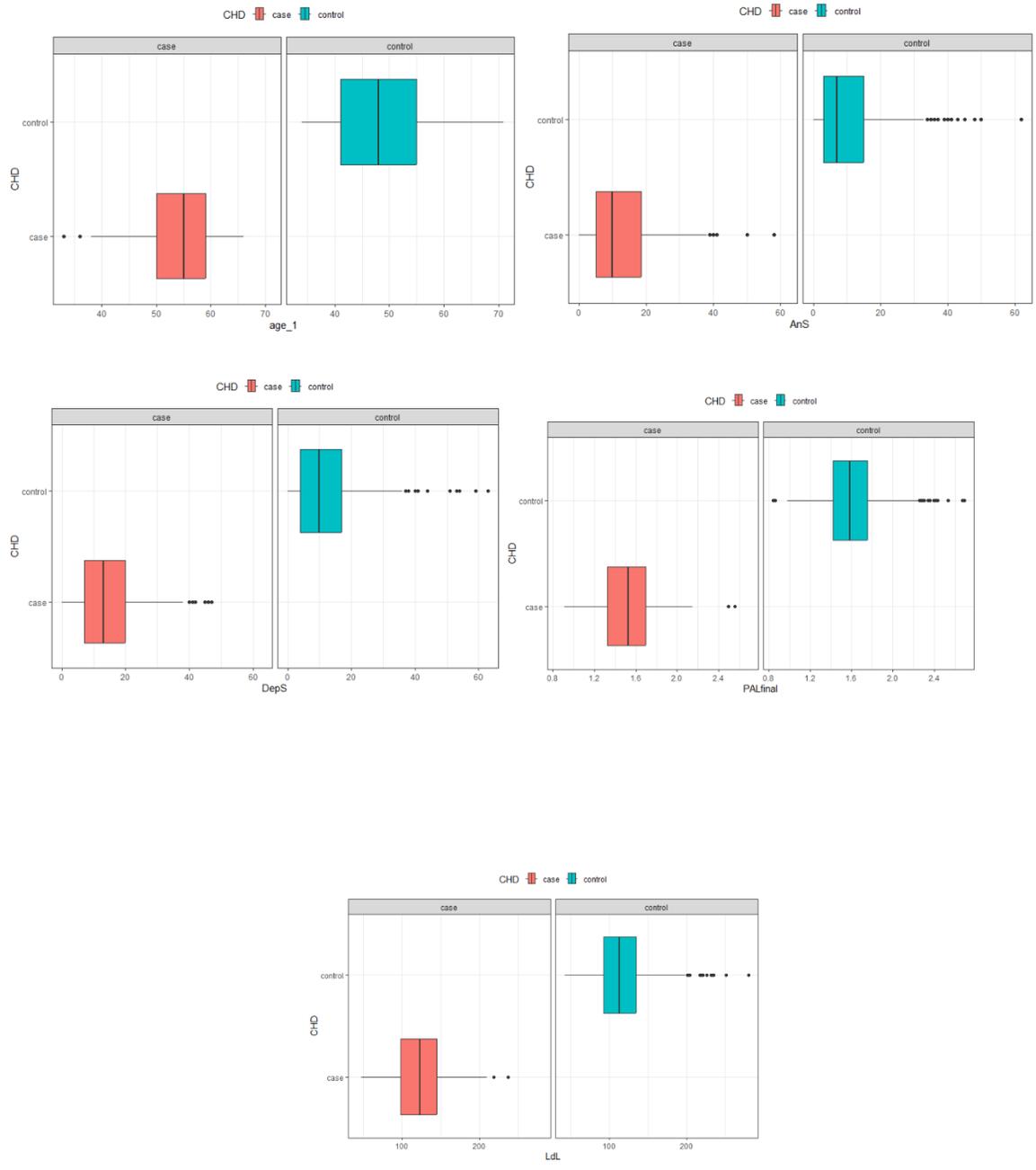

Fig 3: The conditional Box plot for important variables

The results of the classification tree show that age was the most important predictor of coronary artery disease, as the tree model placed it at the root. The cut-off point for the CART algorithm is 46 years old. People under the age of 46 are more likely to develop coronary artery disease if they have a history of stroke. People who are at least 46 years old and don't have a history of hypertension but had myocardial infarction are more likely to have coronary artery disease. Also, people who are at least 63 years old and have hypertension are at risk (as shown in Fig 1). According to the results of the CART algorithm, some of the most important rules are as follows:

- The risk of coronary artery disease in people who are younger than 46 with a history of myocardial infarction is one.
- The risk of coronary artery disease in people at least 46 years of age who don't have a history of hypertension but have a history of myocardial infarction is 0.89.
- The risk of coronary artery disease in people at least 63 years of age who have a history of hypertension is 0.81.
- The risk of coronary artery disease in people aged 46-62.99 who have a history of hypertension, depression score above 11, and physical activity level greater than 1.5, is 0.69.
- The risk of coronary artery disease in people aged 46-62.99 who have a history of hypertension, a depression score above 11, an physical activity level of less than 1.5, and a BMI of less than 28, is 0.67.

The root of the tree is the same as the CART model, in the CHAID algorithm. The suggested cut-off point for this model is 47 years of age. Also, in this model, age, history of hypertension, and history of myocardial infarction are the most important causes of coronary artery disease. Also, some of the results obtained by the CHAID algorithm are as follows:

- The number of people aged up to 47 years who have a history of myocardial infarction is 6.
- The number of people at least 47 years old who have a history of hypertension is 144.

- The number of people at least 47 years of age who don't have a history of hypertension and myocardial infarction is 246.

The root of the tree is the same as the CART and CHIAD models, in the REP algorithm. The results of the REP algorithm show that among the 29 mentioned factors, age, and history of myocardial infarction, HDL level, and depression score physical activity level, educational background, smoking history, blood pressure history, cholesterol level, BMI, anxiety score, depression score, and Brother's heart disease history are more influential in this algorithm. For example, those who are aged up to 43.5 years and have a history of myocardial infarction are more at risk for coronary artery disease.

J48 and NB algorithms identified almost all independent variables in a different order. Myocardial infarction history and Diabetic were identified as the root of the J48 and NB algorithms. The values of sensitivity, specificity, and area under the performance curve were reported by the RF algorithm. This algorithm makes decisions based on collective wisdom. The disadvantage of this algorithm is the lack of identification of important factors and its advantage is the use of collective wisdom in the final tree report. The results of the implementation of these algorithms are shown in Table 3.

| Algorithm | Sensitivity | Specificity | Accuracy | AUC Roc |
| --- | --- | --- | --- | --- |
| CART | 0.79 | 0.45 | 0.75 | 0.73 |
| CHAID | 0.85 | 0.60 | 0.80 | - |
| J48 | 0.80 | 0.46 | 0.73 | 0.59 |
| NB | 0.81 | 0.50 | 0.75 | 0.70 |
| RF | 0.77 | 0.58 | 0.76 | 0.77 |
| REP | 0.79 | 0.49 | 0.74 | 0.70 |

Table 3: Sensitivity, Specificity, Accuracy and AUC Roc of algorithms

**Discussion:**

The decision tree, a powerful data mining method, plays a crucial role in extracting knowledge from data, with variable selection and data processing significantly influencing knowledge discovery(23). This method is particularly apt for problems with a single answer within a class, such as determining the presence of breast cancer(24). Organizing and presenting the knowledge gained from tree construction is essential, offering insights into factors impacting individuals with coronary artery disease (CAD). Identifying a high-risk group through efficient programs can enhance screening efficacy, especially as high-risk populations are more likely to engage in screening when test results indicate a positive outcome(25). Previous studies underscore the efficacy of various methodologies in addressing cardiovascular issues. Liu et al. (26).
utilized AIEchoDx, a deep learning framework, to distinguish common cardiovascular diseases with accuracy comparable to cardiologist consensus. Johri et al. (27) employed clusters from carotid ultrasound and other variables to assess coronary artery stenosis. Ding et al.(28) Developed a deep learning model for coronary heart disease assessment through retinal fundus photographs. Oh et al(29). focused on machine learning classifiers, highlighting hypertension, age, and sex as significant cardiovascular disease risk factors.Further studies employed diverse algorithms for risk assessment. Karaolis et al(30) used the C4.5 algorithm, emphasizing age, sex, and cholesterol as key risk factors. Salehi et al(31). utilized survival competing risks analysis, identifying glucose, diabetes, body mass index, and age as impactful on CAD. Mohammadpour et al. (32). employed neural networks, noting age, total cholesterol, and hypertension as shared risk factors. Sabagh Gol (33). and Kurt et al.(34) Used the C4.5 algorithm and logistic regression, respectively, identifying age as a common risk factor. Mobely et al. (35) highlighted the value of artificial neural networks in predicting coronary artery stenosis. Soni et al. favored decision trees over Bayesian and clustering models for heart disease prediction. (36). Neshati tanha et al.(37) utilized the CHAID algorithm and logistic regression, emphasizing age as a key factor.Tsien et al.(38) compared a classification tree and logistic regression, finding age and a family history of myocardial infarction as shared risk factors.

Purusothaman and Krishnakumari(39) reported a hybrid approach with 0.96 accuracy for heart disease risk prediction. In conclusion, a range of methodologies and algorithms demonstrate effectiveness in cardiovascular risk assessment, with shared risk factors including age, sex, and cholesterol. The diverse approaches provide valuable insights into identifying and managing coronary artery disease.

**Comparison of studies by model:**

In Nashati et al.'s study(37) the logistic model outperformed the CHAID algorithm. Conversely, Tsien et al. (38)favored the decision tree model over the logistic model. Soni et al. (36) proposed the decision tree model in comparison with the Bayesian model and clustering. Sabagh Gol(33) and Karaolis et alxplored the C4.5 algorithm. Salehi et al. (20) considered survival competing risks analysis with a decision tree. In the research of Kurt et al.(34)by comparing the logistic model, decision tree and neural network, neural network model was proposed. In the research of

Purusothaman and Krishnakumari (39)by comparing decision tree model, associative rules, hybrid approach, Bayesian network, supported vector machine and neural network, hybrid approach, is proposed. In the research of Mohammadpour et al (32)emphasized the importance of selecting input variables. In a research by Mobley et al. (35), the neural network model was found to be suitable for identifying heart patients who did not need angiography.

**Comparison of researches based on the findings:**

In statistical studies, it is possible to compare models with each other when the same input variables are included in the study, but in the above studies only a few of the input variables are identical in a few cases. In this study, a classification tree model with CART, J48, NB, REP, RF and CAHID algorithms was used to identify prognostic factors for coronary artery disease.

In all models, the same variables were entered and the result as follows: The most important prognostic factors in coronary artery disease were determined in CART and CHAID models as follows: Age, history of myocardial infarction, history of hypertension, depression score, physical activity level, body mass index and Age, history of myocardial infarction, hypertension. The factors identified by the NB, REP Tree and J48 algorithms are very numerous. Diabetic in NB, history of myocardial infarction in REP Tree and age in CART, CHAID and J48 was selected as root.

Respectively, CHAID has the highest accuracy, sensitivity, and specificity 0.80, 0.85, and 0.60 among other algorithms. These results indicate that factors such as age, history of myocardial infarction, history of hypertension, history of diabetes, cholesterol levels, blood lipids, heart disease in the family are the most important factors among the other models that identified by at least two models of the above algorithms. It has also been identified as a risk factor in some previous researches.

Although it is easier and more practical to infer and obtain a diagnostic method based on the CART algorithm, if the physician aims to achieve a model with higher accuracy and sensitivity, use the CHAID algorithm and if the area under the performance characteristic curve is important, it's better use random forest model.

The NB, REP, and J48 algorithms are almost identical in terms of accuracy and sensitivity. The area under the curve of performance characteristics of the NB and REP models is the same. Both the REP and NB algorithms showed very close results.

• Acknowledgements: This article is part of a Master's thesis in Biostatistics and the authors are grateful to Golestan University of Medical Sciences and Mashhad University of Medical Sciences.